\pdfoutput=1

\documentclass[11pt]{article}

\usepackage[review]{acl}

\usepackage{times}
\usepackage{latexsym}
\usepackage{graphicx} 
\usepackage{multirow}
\usepackage{booktabs}

\usepackage[T1]{fontenc}

\usepackage[utf8]{inputenc}

\usepackage{microtype}

\title{Clozer: Adaptable Data Augmentation for Cloze-style Reading Comprehension}

\author{Holy Lovenia\thanks{\hspace*{0.5em}The authors contributed equally to this work.}, \ Bryan Wilie$^{*}$, \ Willy Chung$^{*}$, \ Min Zeng$^{*}$, \\ \textbf{Samuel Cahyawijaya}, \ \textbf{Su Dan}, \ \textbf{Pascale Fung} \\
    Center for Artificial Intelligence Research (CAiRE) \\
    The Hong Kong University of Science and Technology \\
    \texttt{(hlovenia, bwilie, whcchung, min.zeng)@connect.ust.hk}}

\begin{document}

\nolinenumbers

\maketitle

\begin{abstract}





Task-adaptive pre-training (TAPT) alleviates the lack of labelled data and provides performance lift by adapting unlabelled data to downstream task. Unfortunately, existing adaptations mainly involve deterministic rules that cannot generalize well. Here, we propose Clozer, a sequence-tagging based cloze answer extraction method used in TAPT that is extendable for adaptation on any cloze-style machine reading comprehension (MRC) downstream tasks. We experiment on multiple-choice cloze-style MRC tasks, and show that Clozer performs significantly better compared to the oracle and state-of-the-art in escalating TAPT effectiveness in lifting model performance, and prove that Clozer is able to recognize the gold answers independently of any heuristics.

\end{abstract}




\section{Introduction}

Endowing machines with the proficiency to read, understand, and reason from unstructured text information is an ongoing aspiration in natural language processing. This aim raises a notable research focus: machine reading comprehension (MRC). Given a question, the goal of MRC is to infer the correct answer based on important cues gathered through understanding relevant context passage. MRC tasks vary in structure, depending on their question construction (e.g., cloze-style) and answer type (e.g., multiple-choice)~\cite{zeng2020survey}.

\begin{figure}[!t]
    \centering
    \includegraphics[width=0.95\linewidth, trim={17cm 8.25cm 17cm 8.25cm}, clip]{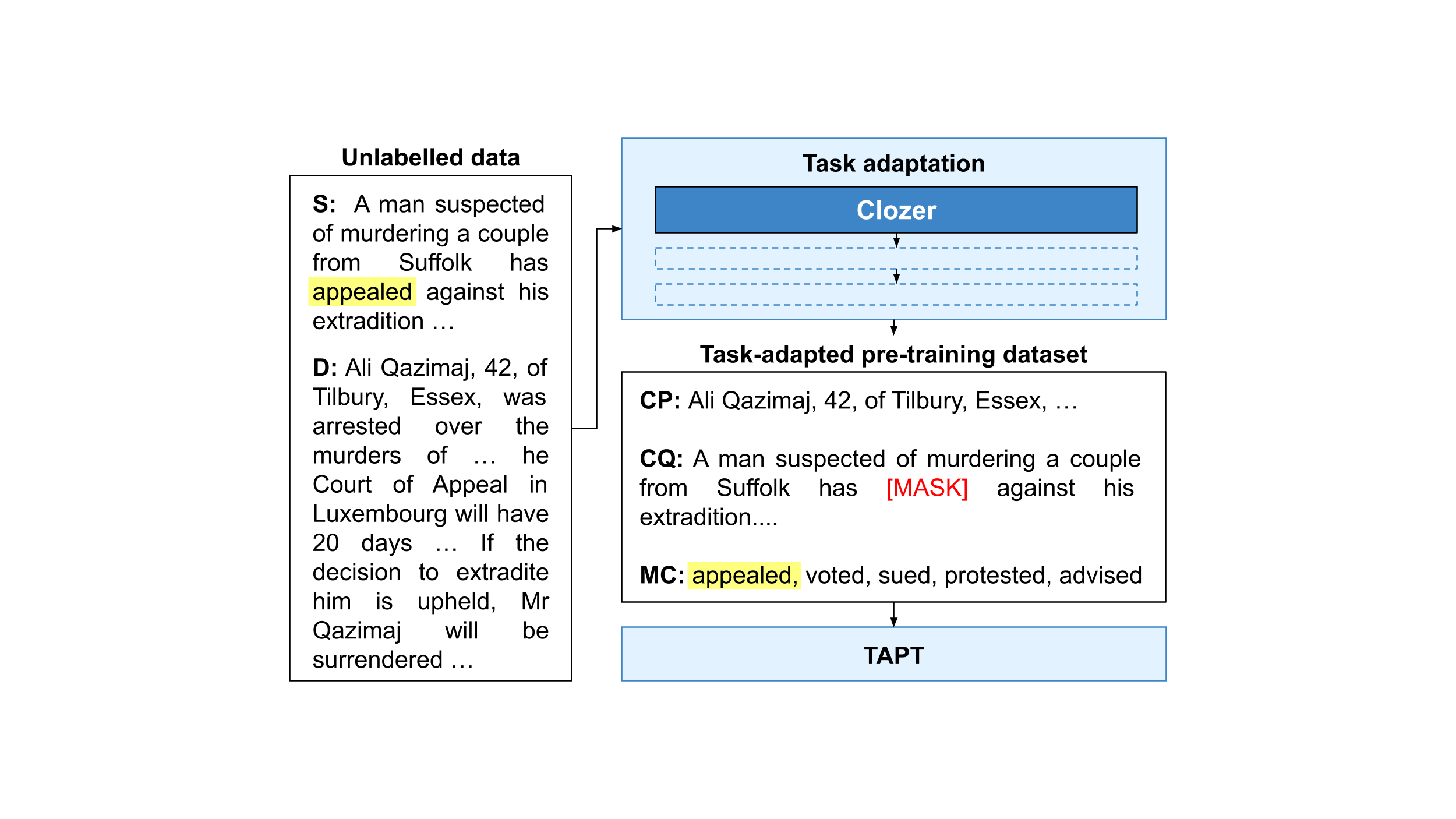}
    \caption{Clozer extracts an answer for TAPT}
    \label{fig:intro}
\end{figure}

Various methods using large pre-trained language models (LMs) have been proposed in MRC tasks. In recent years, adaptation methods such as task adaptive pre-training (TAPT) have been widely adopted for MRC tasks~\cite{xie-etal-2021-zjuklab,wang-etal-2021-pingan,glass-etal-2020-span}. TAPT uses in-domain unlabelled data of the downstream task to generate a synthetic pre-training dataset adapted to the downstream task through certain data augmentation methods, depending on the downstream task in use. For multiple-choice cloze-style MRC, data augmentation often involves two steps: 1) answer extraction or selection and 2) pseudo-answer generation (Figure \ref{fig:intro}). Both steps have been adopted in several studies with varying implementations~\cite{welbl2017crowdsourcing,onishi2016did,yang2020generative}. One notable work presents TA-MAMC~\cite{gururangan-etal-2020-dont}, which achieves state-of-the-art performance by adopting the TAPT framework. However, this method relies heavily on the downstream task's heuristics in the answer selection step, which hinders its applicability to other multiple-choice cloze-style MRC tasks.

\begin{figure*}[!t]
    \centering
    \includegraphics[width=0.95\linewidth,trim={1.5cm 13.75cm 1.5cm 10cm},clip]{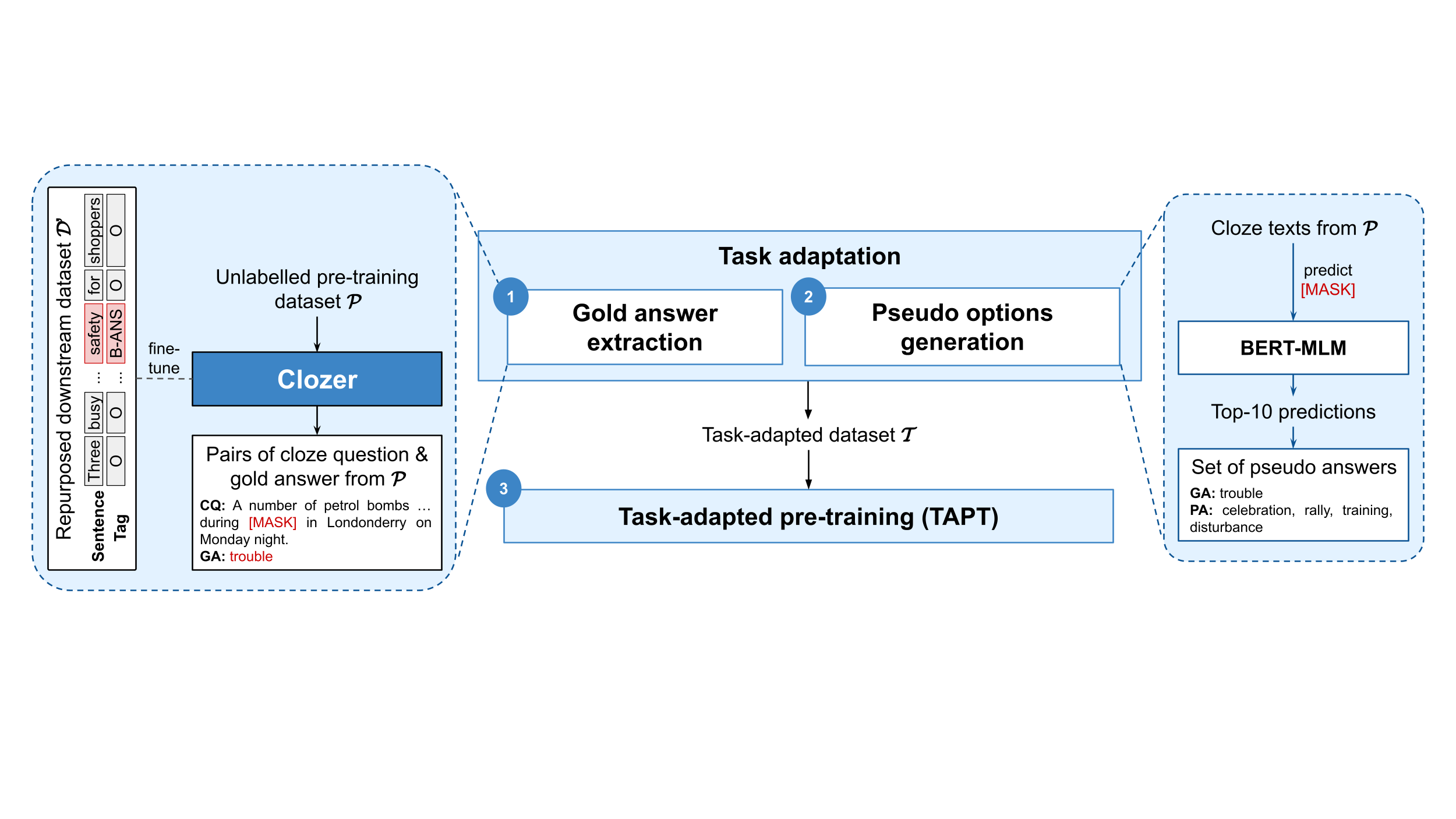}
    \caption{Method pipeline for Clozer-based TAPT}
    \label{fig:system-overview}
\end{figure*}

In this paper, we take a step towards generalized synthetic pre-training dataset construction, to use TAPT to solve multiple-choice cloze-style MRC. We propose Clozer, a cloze answer extraction based on sequence tagging developed independently of pre-defined rules to improve the generalizability of the TAPT method for the cloze-style MRC tasks. Clozer learns the intrinsic pattern of the downstream task dataset and acts as an answer extractor for the unlabelled data (Figure \ref{fig:intro}). To adapt to the downstream task, the extractions are grouped with several other options to form a triplet of \{\emph{context passage, cloze question, multiple-choice options}\}, following the standard multiple-choice cloze-style MRC task format, as a synthetic sample for the second pre-training phase. We conduct our experiments on two downstream tasks. Our experimental results show that employing Clozer in TAPT provides a substantial performance boost, while being generally applicable for both multiple-choice MRC tasks we experiment on.

Our contributions are as follows:
1) to the best of our knowledge, we are the first to introduce an automatic generalizable cloze answer extraction method to support a generalized TAPT method for multiple-choice cloze-style MRC tasks;
2) we show that Clozer significantly outperforms all other baselines on two multiple-choice cloze-style MRC tasks without relying on any task-specific heuristics; and
3) we present further analysis to explain the effectiveness and efficiency of our Clozer and provide insight on how to improve its generalizability.

\section{Related Work}

\paragraph{Task-adaptive pre-training}

\citet{howard2018universal} propose Universal Language Model Finetuning (ULMFiT), which pre-trains an LM on a large general-domain corpus and fine-tunes it on the target task. Second-phase pre-training has been used to improve the performance of an LM for certain downstream tasks such as text classification \cite{DBLP:journals/corr/abs-1905-05583}. Studies on TAPT \cite{gururangan-etal-2020-dont,pruksachatkun2020intermediate} prove that the performance boost it obtains can be on par with domain-adaptive pre-training, with the benefit of using a much smaller but relevant corpus. TAPT has proved effective in many downstream tasks such as abstractive summarization ~\cite{yu2021adaptsum} and dialogue systems~\cite{zhang2021hybrid}.

\paragraph{Answer extraction}
\citet{tan2018s} develop an extraction-then-synthesis framework to synthesize answers from extraction results. Specifically, the answer extraction model is first employed to predict the most important sub-spans from the passage, then the answer synthesis model takes the sub-spans as additional features along with the question and passage to further elaborate the final answers. \citet{xiong2016dynamic} introduce the Dynamic Coattention Network (DCN) for a question-answering task, which learns the co-dependent representations of the question and the passages. 
\citet{seo2016bidirectional} introduce the Bi-Directional Attention Flow (BIDAF) network to match the question and passages. It uses the BIDAF mechanism to get a query-aware context representation without early summarization.


\paragraph{Sequence tagging}

Sequence tagging is utilized to assign a label for each token (i.e., word) in a sequence. While it's commonly applied for tasks like named entity recognition (NER), part-of-speech (POS) tagging, and text chunking, \citet{yao2013answer, wilie-etal-2020-indonlu} prove that it is feasible to use this approach to construct cloze questions by extracting an answer span from a complete sentence. \citet{yao2013answer} cast answer extraction as an answer sequence-tagging task, utilizing a linear-chain conditional random field (CRF) with tree edit distance (TED) and traditional contextual features.

\section{Methodology}
\label{sec:methodology}

Our method follows the pipeline described in Figure \ref{fig:system-overview}. We follow TAPT's objective, in which a model learns on a small task-relevant set of data instead of doing another round of masked language modeling (MLM) for pre-training. Utilizing Clozer, we adapt a large unlabelled pre-training dataset based on the downstream task, which could be any multiple-choice cloze-style MRC task.

We define the pre-training dataset $\mathcal{P} = \{(d^\mathcal{P}_i, s^\mathcal{P}_i)\}^n_{i=1}$ with $d^\mathcal{P}_i$ as a document and $s^\mathcal{P}_i$ as a summary or a single sentence related to the passage $d^\mathcal{P}_i$. $\mathcal{P}$ could be any unlabelled data of document and sentence pairs, e.g., headline-content of news, title-body of articles, and synopsis-narration of stories. Through the task adaptation, we reconstruct $\mathcal{P}$ into a synthetic cloze-style MRC task, where the resulting task-adapted pre-training dataset is represented by $\mathcal{T} = \{(c^\mathcal{T}_i, q^\mathcal{T}_i, o^\mathcal{T}_{i}, l^\mathcal{T}_i)\}^m_{i=1}$. It follows the structure of the downstream task dataset $\mathcal{D} = \{(c^\mathcal{D}_i, q^\mathcal{D}_i, o^\mathcal{D}_{i}, l^\mathcal{D}_i)\}^m_{i=1}$, where $c^\mathcal{D}_i$ is a context passage, $q^\mathcal{D}_i$ is a cloze question, $o^\mathcal{D}_{i} \in {o_1,\dots,o_k}$ is a set of multiple-choice options, and $l^\mathcal{D}_i$ is the gold answer's index as the correct label.

We split the task adaptation into 1) gold answer extraction and 2) pseudo options generation, which are explained in \S\ref{sec:gold-answer-extraction} and \S\ref{sec:pseudo-options-generation} respectively. Afterwards, we employ TAPT using the task-adapted dataset $\mathcal{T}$, the details of which are provided in \S\ref{sec:tapt}.


\subsection{Gold answer extraction}
\label{sec:gold-answer-extraction}

Gold answer extraction (GAE) represents the pre-training dataset's summary as a cloze question by taking out a gold answer, which depends on the downstream task's notion of what is a correct answer. We tackle this problem by utilizing Clozer to learn from the downstream task and identify the appropriate gold answers by sequence tagging. First, we repurpose the cloze questions and gold answers in the downstream task as a token classification dataset. We use the tag \texttt{B-ANS} for the gold answer and the tag \texttt{O} for other words in the cloze question.

Afterwards, we fine-tune Clozer on this repurposed dataset so it can learn and approximate the downstream task's pattern of determining the gold answers. It is worth mentioning that, due to its independence from any heuristic rules, our Clozer method is not constrained to a single specific interpretation of gold answers. It can be adapted to extract any type of cloze answers (e.g., abstract meaning) depending on the downstream task dataset. We next use Clozer to predict the pre-training dataset's summaries and extract the gold answers. We replace the gold answers in the summaries with the \texttt{[MASK]} token to form cloze questions and pass the questions on to the next step. We drop candidates with zero or more than one gold answer.

\subsection{Pseudo options generation}
\label{sec:pseudo-options-generation}

Pseudo answer generation (POG) employs a pre-trained masked LM to predict the \texttt{[MASK]} token. For each cloze question, we obtain the model's top predictions and filter out the ones that are incomplete or too similar to the gold answer. We randomly pick $k$ predictions as pseudo options. We discard data samples with fewer than $k$ remaining predictions.
After this step, each pre-training dataset sample consists of a context paragraph, a cloze question, a gold answer, and four pseudo options.
Following the downstream task dataset structure, we recast the gold answer and pseudo options as $\{o_1, o_2, ..., o_k\}$ in random order. The gold answer's option index becomes the label. In cases beyond the scope of this work where multiple-choice is not required by the cloze task, POG is skipped.

\subsection{Task-adaptive pre-training}
\label{sec:tapt}

We feed the task-adapted dataset to a pre-trained multiple-choice classification model for TAPT. The final step is to fine-tune the model on the downstream task and evaluate it. To see how Clozer performs against other available methods, we present the results of three baselines, where we employ a directly fine-tuned model, TA-MAMC, and an oracle in place of Clozer in the GAE step. The baselines will be further explained in \S\ref{sec:baseline}.

\section{Experiment}
\label{sec:exp-settings}

\paragraph{Dataset}
\label{sec:dataset}

As explained in \S\ref{sec:methodology}, the methodology requires the usage of a pre-training dataset and a downstream task. In the experiment, we apply Clozer for the TAPT method on two downstream tasks separately. Both are multiple-choice cloze-style MRC tasks and are obtained from the subtask 1 and subtask 2 of ReCAM \cite{zheng-etal-2021-semeval}. Given a context passage and multiple choice options, the appropriate gold answer must be derived to complete a cloze question. The first task defines its gold answers as imperceptible concepts, while the second defines them as hypernyms. For the pre-training dataset $\mathcal{P}$, we use XSUM \cite{narayan-etal-2018-dont}, an abstractive news summarization dataset.

\paragraph{Baseline}
\label{sec:baseline}

To see how Clozer-based TAPT performs against other methods, we employ three baselines for the experiment: \textbf{1) direct fine-tuning}, where a pre-trained multiple-choice model applies no TAPT and is immediately fine-tuned on the downstream task; \textbf{2) TA-MAMC}, which selects gold answers by emulating the POS-tag distribution of the downstream task's training data; and \textbf{3) oracle}, whose answer selection is built upon heuristic rules specific to each downstream task.

The oracle utilizes a psycholinguistic database of abstract words \cite{coltheart1981mrc} to select the \emph{imperceptible} concepts as the gold answers in the first task. For the second task, it uses a hypernym hierarchy from WordNet~\cite{CHANGIZI2008214} to determine the gold answers. Both heuristics are chosen because they are used to select the original gold (i.e., correct) answers in the ReCAM dataset creation.

\paragraph{Training and evaluation}
\label{sec:train-eval}

In the GAE, our Clozer is implemented using a pre-trained ELECTRA-base~\cite{clark2020electra}, while for the POG and TAPT, we initialize the model using a pre-trained BERT-base model~\cite{devlin2019bert}. Since only the training set and the development set of both downstream tasks are labelled, we split the original training set with a ratio of 80:20 to form a training set and a validation set. We use the development set as a test set. Accuracy and F1-score are used to assess the methods' performance on the test set.



\section{Results and Analysis}

\begin{table}[t]
\centering
\resizebox{0.9\linewidth}{!}{%
\begin{tabular}{@{}lcccc@{}}
\toprule
\multicolumn{1}{c}{\multirow{2}{*}{\textbf{Approach}}} & \multicolumn{2}{c}{\textbf{ReCAM 1}} & \multicolumn{2}{c}{\textbf{ReCAM 2}} \\ \cmidrule(lr){2-3}\cmidrule(lr){4-5}
\multicolumn{1}{c}{} & \multicolumn{1}{c}{\textbf{Acc}} & \multicolumn{1}{c}{\textbf{F1}} & \multicolumn{1}{c}{\textbf{Acc}} & \multicolumn{1}{c}{\textbf{F1}} \\ \midrule
Direct FT & 64.16\% & 64.15\% & 64.75\% & 64.65\% \\
TA-MAMC$^\dagger$ & 64.99\% & 64.99\% & 67.69\% & 67.68\% \\
Oracle & 65.83\% & 65.80\% & 68.60\% & 68.50\% \\ \midrule
\textbf{Clozer} & \textbf{65.95}\% & \textbf{65.96}\% & \textbf{73.56}\% & \textbf{73.45}\% \\ \bottomrule
\end{tabular}%
} 
\caption{Performance comparison on the test sets of the downstream tasks. \textbf{Bold} marks the best results. $^\dagger$We reproduce this approach based on~\citet{zhang-etal-2021-ta}.}
\label{tab:result}
\end{table}

\subsection{Overall results}

We present our experimental results in Table \ref{tab:result}. Without additional TAPT, the direct fine-tuning method yields the lowest results. In comparison, TA-MAMC, which relies on POS-tag distribution, performs slightly better, and the oracle, which exploits the downstream tasks’ heuristic rules, achieves the best scores among our baselines. Our proposed Clozer method, however, surpasses all baselines in both downstream tasks, by around 2\% for task 1 and 9\% for task 2. While Clozer provides substantial improvements, there is a considerable discrepancy between both performances due to the way the tasks are defined. We further discuss Clozer's performance discrepancy in \S\ref{sec:TAPT_analysis}.

\subsection{Quality of answer extraction methods}

As shown in Table \ref{tab:passage_summary_result}, the oracle, which derives its understanding of the answers from the semantics provided by the heuristic rules, has the fewest data after the GAE step (94k out of 200k), because the heuristic rules it is built upon are deterministic and leave no room for randomness. TA-MAMC's POS-tag distribution approach provides some knowledge of the target's syntax but represents no semantic ties to ReCAM's answers (i.e., imperceptible concepts and hypernyms).

However, TA-MAMC has the benefit of excluding fewer examples than the oracle. Our Clozer finds a middle ground by being more generalizable compared to both baselines, while producing a better answer extraction quality (Table \ref{tab:result}).
Clozer shows superior results with only 5k more data samples in task 2 and with 12k fewer data samples in task 1. This shows that, while the amount of data contributes to the performance lift, the quality of the extracted answers in the synthetic task-adapted dataset is indispensable.

\begin{table}[!t]
\centering
\resizebox{0.95\linewidth}{!}{%
\begin{tabular}{lcccc}
    \toprule
    \multicolumn{1}{c}{\multirow{3}{*}{\textbf{Task adapter}}} & \multicolumn{2}{c}{\textbf{ReCAM 1}} & \multicolumn{2}{c}{\textbf{ReCAM 2}} \\ \cmidrule(lr){2-3}\cmidrule(lr){4-5}
    \multicolumn{1}{c}{} & \multirow{2}{1.3cm}{\centering \textbf{Post-GAE}} & \multirow{2}{1.3cm}{\centering \textbf{Post-POG}} & \multirow{2}{1.3cm}{\centering \textbf{Post-GAE}} & \multirow{2}{1.3cm}{\centering \textbf{Post-POG}} \\
     & & & & \\
    \midrule
    TA-MAMC & 155017 & 47699 & 155858 & 48358 \\
    Oracle & 94954 & 29073 & 75920 & 23520 \\
    \midrule
    \textbf{Clozer} & 120073 & 35073 & 181476 & 53368 \\
    \bottomrule
\end{tabular}%
}
\caption{Number of data samples left after the \textbf{GAE} and \textbf{POG}
for different task-adapter methods.}
\label{tab:passage_summary_result}
\end{table}

\subsection{Clozer's performances on different downstream tasks}
\label{sec:TAPT_analysis}

While TAPT lifts the model performance by 2\% for task 1 and 9\% for task 2, the difference between the tasks is glaring. We argue that this is largely due to the amount of synthetic data left after applying the task adaptation, as shown in Table \ref{tab:passage_summary_result}, with 35k samples left in task 1 and 53k samples in task 2. This shows that the definition of abstractness chosen by ReCAM for gold answers in task 1 is more complex than the definition used by task 2, which causes the answers in task 1 to be harder to grasp by all of the approaches, including our Clozer.

This is coherent as ReCAM defines \emph{imperceptible} concepts in task 1 using a model-based approach, which in turn introduces an innate bias to the definition. This causes identifying answers in task 1 to be conceptually more complex than in task 2, where the answers are simply nouns and verbs derived from a hypernym hierarchy. This is also in line with \citet{zheng-etal-2021-semeval}, who show that the cross-task performance drops significantly more for models trained on task 2 trying to make predictions on task 1, rather than the opposite. Examples of this complexity difference are in Appendix \ref{sec:examples}.




\section{Conclusion}

We have proposed Clozer, an automatic generalizable cloze answer extraction method, to help in synthetic TAPT dataset construction in multiple-choice cloze-style MRC tasks. Performing TAPT with gold answers extracted by our ELECTRA-based Clozer produces stronger models than the baselines in terms of effectiveness (i.e., performance) and efficiency (i.e., the amount of data used in TAPT). Moreover, we also show that the quality of Clozer's extracted answers is higher, despite its independence from the downstream task's heuristics

\section*{Acknowledgement}

This work has been supported by the China NSFC Project (No. NSFC21EG14), School of Engineering PhD Fellowship Award, the Hong Kong University of Science and Technology and PF20-43679 Hong Kong PhD Fellowship Scheme, Research Grant Council, Hong Kong.



\bibliography{anthology,custom}
\bibliographystyle{acl_natbib}

\newpage
\onecolumn
\appendix{}

\setcounter{table}{0}
\renewcommand{\thetable}{A\arabic{table}}
\setcounter{figure}{0}
\renewcommand{\thefigure}{A\arabic{figure}}
\section{Examples of Gold Selection with Clozer}
\label{sec:examples}

\begin{table}[h]
\centering
\begin{tabular}{p{15cm}}
\toprule{}
David Beckham has expressed his \colorbox{yellow}{pride} at helping London win their 2012 olympics bid despite not being \colorbox{cyan}{picked} in great britains football squad. \\
\midrule{}
A 22 year old man arrested on \colorbox{yellow}{suspicion} of murder \colorbox{cyan}{following} the death of Lewis Siddall has been released on bail. \\
\midrule{}
A cow which got into the \colorbox{cyan}{water} at Aberdeen harbour has been shot after a rescue \colorbox{yellow}{effort} failed to coax it ashore. \\
\midrule{}
Streets in Wales are blighted by discarded cigarette butts with 86 of roads \colorbox{cyan}{strewn} with smoking related litter a \colorbox{yellow}{charity} survey shows. \\
\midrule{}
It is officially a \colorbox{cyan}{regeneration} area and dyke house in Hartlepool has newly built \colorbox{yellow}{smart} houses but they are in the minority. \\
\midrule{}
Wales flyhalf Dan Biggar says he is learning to cope with the pressure of \colorbox{cyan}{wearing} the \colorbox{yellow}{famous} number 10 jersey. \\
\bottomrule{}
\end{tabular}
\caption{Examples of gold selections in summaries taken from both dowstream tasks with Clozer. Highlighted in \colorbox{yellow}{yellow} is the gold answer chosen according to the first definition of abstractness, \emph{imperceptibility}, and in \colorbox{cyan}{blue} the answer according to the second definition, \emph{non-specificity} (for hypernyms), in each example.}
\label{tab:data_examples}
\end{table}

For task 1 (ReCAM 1), abstractness follows the definition of imperceptibility, meaning any concept that can’t be perceived directly in the physical world according to a psycholinguistic database \cite{coltheart1981mrc}. Task 2 (ReCAM 2) defines abstractness as non-specificity, representing nouns and verbs relatively high in a hypernym hierarchy \cite{CHANGIZI2008214}. Examples of the difference between both are illustrated in Table \ref{tab:data_examples}.

As discussed in \S\ref{sec:TAPT_analysis}, the abstract concepts chosen for ReCAM 1 are intuitively harder to define compared to the concepts for ReCAM 2, even for humans (\textbf{pride, suspicion} vs \textbf{picked, following}). However, this also shows that without being given any rules, our Clozer still manages to grasp the underlying mechanics originally chosen to extract the abstract words in both tasks.
 
We refer to the original work \cite{zheng-etal-2021-semeval} on building the ReCAM dataset for more details on the reason why those two definitions of abstractness have been chosen.

\end{document}